\pdfoutput=1
\documentclass[letterpaper, 10 pt, conference]{ieeeconf}  % Comment this line out if you need a4paper

\usepackage{algorithm, algpseudocode, graphicx,float, dsfont, amsmath,amssymb,cancel,arydshln,bm, etoolbox}
\usepackage{graphicx,amsmath,amssymb,bm,xcolor,soul}
\usepackage{color, colortbl}
\usepackage{multirow}
\usepackage{multicol}
\usepackage{blindtext}
\usepackage{todonotes}
\usepackage[font=small,labelfont=bf]{caption}
\usepackage{cite}
\usepackage{booktabs}

\graphicspath{ {./figures/} }
\patchcmd\subequations
 {\theparentequation\alph{equation}}
 {\subequationsformat}
 {}{}

\newcommand{\subequationsformat}{\theparentequation.\alph{equation}}

\newcommand{\RNum}[1]{\uppercase\expandafter{\romannumeral #1\relax}}

\IEEEoverridecommandlockouts                              % This command is only needed if 
\title{\LARGE \bf
Scale Disparity of Instances in Interactive Point Cloud Segmentation
}

\author{Chenrui Han$^{1}$, Xuan Yu$^{1}$, Yuxuan Xie$^{1}$, Yili Liu$^{1}$, Sitong Mao$^{2}$, Shunbo Zhou$^{2}$, Rong Xiong$^{1}$, Yue Wang$^{1}$% <-this % stops a space}
\thanks{$^{1}$Chenrui Han, Xuan Yu, Yuxuan Xie, Yili Liu, Rong Xiong and Yue Wang are with Zhejiang University, Hangzhou, Zhejiang, China. Yue Wang is the corresponding author {\tt\footnotesize wangyue@iipc.zju.edu.cn}}%
\thanks{$^{2} $Sitong Mao and Shunbo Zhou are with Huawei Cloud Computing Technologies Co., Ltd., Shenzhen, China.}%
}

\usepackage{etoolbox}
\makeatletter
\patchcmd{\@makecaption}
  {\scshape}
  {}
  {}
  {}
\makeatother

\usepackage{url}
\usepackage[colorlinks,
            allcolors=blue,
            linkcolor=blue,
            anchorcolor=blue,
            citecolor=blue]{hyperref}

\begin{document}
\maketitle
\thispagestyle{plain}
\pagestyle{plain}
%%%%%%%%%%%%%%%%%%%%%%%%%%%%%%%%%%%%%%%%%%%%%%%%%%%%%%%%%%%%%%%%%%%%%%%%%%%%%%%%
\begin{abstract}

Interactive point cloud segmentation has become a pivotal task for understanding 3D scenes, enabling users to guide segmentation models with simple interactions such as clicks, therefore significantly reducing the effort required to tailor models to diverse scenarios and new categories. However, in the realm of interactive segmentation, the meaning of instance diverges from that in instance segmentation, because users might desire to segment instances of both thing and stuff categories that vary greatly in scale. Existing methods have focused on thing categories, neglecting the segmentation of stuff categories and the difficulties arising from scale disparity. To bridge this gap, we propose ClickFormer, an innovative interactive point cloud segmentation model that accurately segments instances of both thing and stuff categories. We propose a query augmentation module to augment click queries by a global query sampling strategy, thus maintaining consistent performance across different instance scales. Additionally, we employ global attention in the query-voxel transformer to mitigate the risk of generating false positives, along with several other network structure improvements to further enhance the model's segmentation performance. Experiments demonstrate that ClickFormer outperforms existing interactive point cloud segmentation methods across both indoor and outdoor datasets, providing more accurate segmentation results with fewer user clicks in an open-world setting. Project page: \url{https://sites.google.com/view/clickformer/}

\end{abstract}

%%%%%%%%%%%%%%%%%%%%%%%%%%%%%%%%%%%%%%%%%%%%%%%%%%%%%%%%%%%%%%%%%%%%%%%%%%%%%%%%
\section{Introduction}
\label{sec:intro}

Interactive point cloud segmentation is an emerging task in the understanding of 3D scenes. It allows human users to interact with the segmentation model through low-cost prompts like clicks to facilitate the model's adaptation to diverse scenarios and novel categories with minimal effort. However, in the realm of interactive segmentation, the meaning of instance diverges from the predefined thing categories in instance segmentation. As a crucial component of point cloud semantics, users may desire to achieve segmentation results of a stuff category through the same interactive mode, in other words, to \emph{segment stuff as an instance}. This means that the instance in interactive segmentation can be larger than a wall or smaller than a cup. The scale disparity of instances presents a significant challenge.

\begin{figure} [t]
    \centering
    \includegraphics[width=\linewidth]{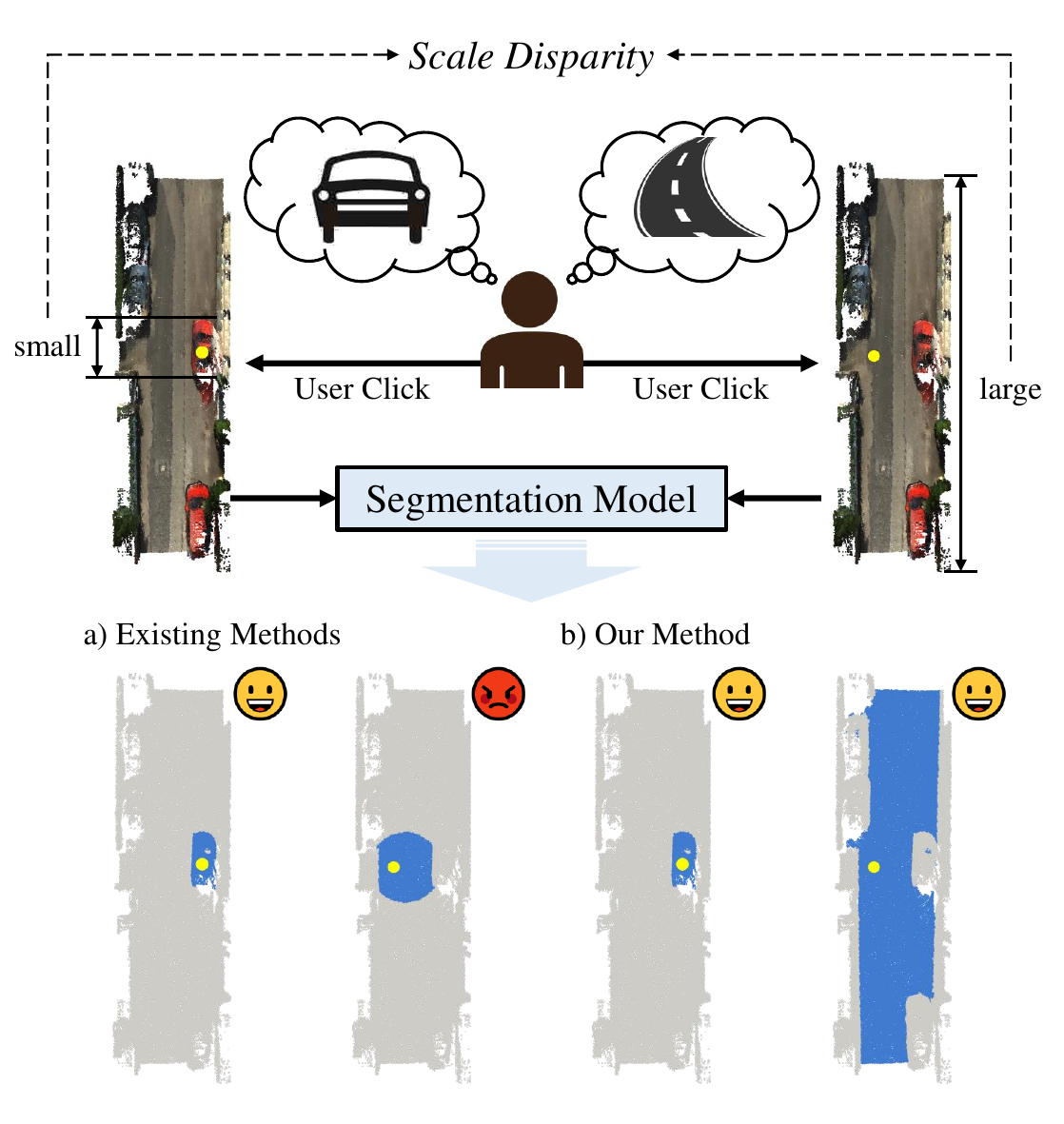}
    \caption{\textbf{Scale Disparity in Interactive Segmentation. } Existing methods suffer from scale disparity when segmenting stuff categories. \vspace{-3.5mm}
    }
    \label{fig:teasing}
    \vspace{-4mm}
\end{figure}

Most existing methods for interactive point cloud segmentation\cite{sun2023click, kontogianni2023interactive} focus primarily on instances of thing categories, similar to instance segmentation setting, thereby overlooking the challenges posed by the scale disparity of instances. Yue et al.\cite{yue2023agile3d} introduced a novel problem setting that segments multiple instances simultaneously, yet they did not take into consideration the segmentation of stuff categories. Thus, the question arises: \emph{How to address the scale disparity of instances in point cloud segmentation models?}

Interactive segmentation methods based on sparse convolutional UNet\cite{kontogianni2023interactive, sun2023click} struggle with a common issue inherent to CNN-based segmentation methods, the limited receptive field, leading to poor performance in segmenting large-scale instances. The transformer-based method\cite{yue2023agile3d} might be a potential solution to this problem. However, it only use user clicks as queries, akin to 2D interactive segmentation methods. The number of queries is significantly less than that typically used in transformer-based general point cloud segmentation methods, resulting in a lower coverage of the point cloud by the queries' receptive fields. This limitation could greatly impact its ability to segment large-scale instances as well. Based on the considerations mentioned above, we believe that the quantity of queries and the scope of attention could be key to solving the scale disparity.

In this paper, inspired by transformer-based general point cloud segmentation methods, we introduce a query augmentation module sampling additional queries to address the scale disparity of instances. The query augmentation module prevents the attention scopes of queries from drastically changing with the scale of the instances, thereby improving model performance, especially in segmenting large-scale instances. However, The augmented queries make it easier for the model to generate false positives when segmenting small-scale instances. To tackle this issue, we employ global attention instead of the common used local attention in point cloud transformers to facilitate queries in exchanging global information. It helps to prevent augmentation queries from generating false positives due to focusing solely on local neighborhoods. Considering that the number of queries used by our model is significantly lower than that in general point cloud segmentation transformers, the increased computational cost associated with global attention is manageable. In summary, our main contributions are as follows:
\begin{itemize}
    \item We propose an interactive point cloud segmentation method ClickFormer, which can more accurately segment both thing and stuff categories.
    \item We propose the query augmentation module to address scale disparity and improve the segmentation of large-scale instances.
    \item We adopt global attention along with several other network structure improvements to further enhance the model's segmentation performance.
    \item Our method achieves superior performance than existing interactive segmentation methods on both indoor and outdoor point cloud datasets.
\end{itemize}

%%%%%%%%%%%%%%%%%%%%%%%%%%%%%%%%%%%%%%%%%%%%%%%%%%%%%%%%%%%%%%%%%%%%%%%%%%%%%%%%
\section{Related Works}
\label{sec:rw}
\subsection{Point Cloud Segmentation}
3D point cloud segmentation methods, like other fully supervised methods, face a pressing challenge: the contradiction between the demand for a substantial volume of fully annotated data and the high annotation costs for point clouds\cite{gao2020we}. To address the issue of data hunger, several weakly and semi-supervised methods have been proposed. Xu et al.\cite{xu2020weakly} trained on point cloud data that was only partially annotated, introducing spatial and color smoothness as additional supervisory constraints. Unal et al.\cite{unal2022scribble} proposed a method trained on scribble-annotated data, employing a teacher model to generate pseudo-labels, thus supplementing the supervisory data for unannotated points. While achieving impressive results, their focus has primarily been on reaching the performance of fully supervised methods with less annotated data, rather than on enhancing segmentation performance. Consequently, these methods still fail to address the challenge of generalizing point cloud segmentation to different scenes and unseen categories in the open world.
\subsection{Interactive Image Segmentation}
Interactive image segmentation has garnered considerable research interest. In recent developments, the mode of interaction has largely centered around human clicks\cite{sofiiuk2022reviving, chen2022focalclick, liu2023simpleclick, du2023efficient}. Sofiiuk et al. \cite{sofiiuk2022reviving} proposed a network that iteratively refines segmentation masks by incorporating existing masks and encoded clicks. Liu et al. \cite{liu2023simpleclick} utilized a plain ViT as the segmentation backbone to processes images and click inputs through patch embedding, which is then processed through a lightweight MLP decoder to generate the segmentation mask. Kirillov et al.\cite{kirillov2023segment} developed a model that supports encoding both clicks and language as prompt tokens for transformer input, enhancing the segmentation performance by providing richer intent cues. Unfortunately, due to the inherent differences in data distribution, the straightforward transfer of 2D methods to the 3D domain falls significantly short of achieving the original effects.

\subsection{Interactive Point Cloud Segmentation}

Compared to point cloud segmentation and interactive image segmentation, research on interactive point cloud segmentation is relatively scarce. In the methods developed by researchers such as Shen et al.\cite{shen2020interactive}, Zhi et al.\cite{zhi2022ilabel}, and Goel et al.\cite{goel2023interactive}, the user's prompt clicks are limited to interactions with RGB images. This limitation confines the applicability of these methods to 3D scenes that include imagery and have a fixed camera pose, rendering them incapable of segmenting point cloud inputs from non-camera sensors (e.g., LiDAR). In interactive segmentation methods directly applied to 3D point clouds, Sun et al.\cite{sun2023click} encoded user clicks as an additional attribute of the point cloud, extracting features from the point cloud through MLPs and processing segmentation masks. Kontogianni et al.\cite{kontogianni2023interactive}, on the other hand, encoded user clicks into a 3D volume based on the spatial relationships within the point cloud, obtaining segmentation results through a 3D UNet based on sparse convolutions. However, suffering from scale disparity, these methods all fail to segment large-scale instances of stuff categories.

%%%%%%%%%%%%%%%%%%%%%%%%%%%%%%%%%%%%%%%%%%%%%%%%%%%%%%%%%%%%%%%%%%%
\section{Methodology}
\label{sec:method}

\begin{figure*} [t]
    \centering
    \includegraphics[width=\linewidth]{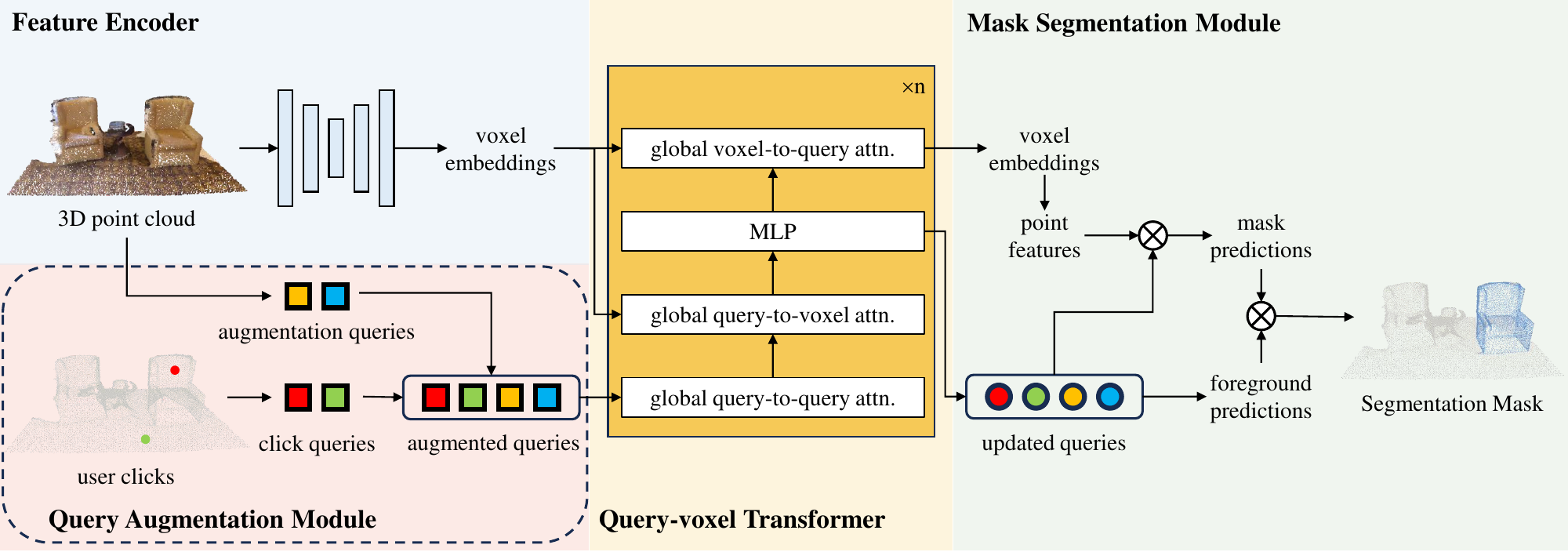}
    \caption{\textbf{Network Architecture.} The overall network consists of 1) a feature encoder, 2) a query augmentation module and 3) a mask decoder composed of a query-voxel transformer and a mask segmentation module. \vspace{-3.5mm}
    }
    \label{fig:network}
    \vspace{-4mm}
\end{figure*}

\subsection{Overview}

\textbf{Problem Definition.} Consider a 3D point cloud $P \in \mathbb{R}^{N \times C}$, where $N$ is the number of points in the point cloud and $C$ is the feature dimension of each point. Then the user provides an arbitrary number of clicks, denoted as $S=\{s_1, s_2, ..., s_t\}$. User clicks can be considered as sparse labels for point clouds, with each one containing two parts of information $s_t=(pos_t, sgn_t)$, where $pos_t\in \mathbb{R}^3$ is the 3D coordinate of the click, and $sgn_t\in \{+1,-1\}$ indicates whether it is a positive click or a negative click. A positive click denotes the point is on the desired mask, while a negative click denotes it belongs to the background. Given the 3D point cloud $P$ and user clicks $S$, the model's objective is to output a binary segmentation mask $M \in \mathds{1}^{N \times 1}$, where $1$ represents the foreground and $0$ represents the background.

\textbf{Network Architecture.} The overall architecture of ClickFormer is illustrated in Fig \ref{fig:network}, consisting of the following three components: 1) a feature encoder that encodes the input point cloud into voxel features, 2) a query augmentation module that encodes user clicks and adopts a global sampling strategy, 3) a mask decoder composed of a query-voxel transformer that allows bidirectional updates between query and voxel embeddings, and a mask segmentation module that ultimately generates the desired mask based on the query and voxel embeddings.

\subsection{Feature Encoder}

In general, the feature encoder can be any network that takes a point cloud $P \in \mathbb{R}^{N \times C}$ as input and outputs voxel features $F \in \mathbb{R}^{N' \times D}$, where $N'$ is the number of encoded voxels, and $D$ is the voxel feature dimension. Considering the pre-training performance and the suitable scene size of the network, we select GD-MAE\cite{yang2023gd} as the feature encoder for ClickFormer in large-scale street scenes, while MinkowskiUNet\cite{choy20194d} as the feature encoder for small-scale indoor scenes.

\subsection{Query Augmentation Module}

\textbf{Click Queries.} Unlike previous works that chose to encode all clicks into a click map as an additional attribute of the point cloud, we encode each click as an independent query. It not only allows an arbitrary number of clicks but also maintains scalability to more possible forms of prompts.

\textbf{Augmentation Queries.} Besides user clicks, we employ a global query sampling strategy to select additional points as augmentation queries within the point cloud. To ensure that the queries are most evenly distributed in space, we utilize FPS (Farthest Point Sampling) to pick augmentation query points from the global point cloud. Due to the uniform spatial distribution of augmentation queries, larger instance scales lead to more queries participating in generating the desired mask. This ensures that the attention scope of each query is independent of the scale of the instance to be segmented, without increasing the number of user clicks and thus avoiding additional manual costs. Consequently, this eliminates the impact of scale disparity on the interactive segmentation model.

\begin{figure} [t]
    \centering
    \includegraphics[width=\linewidth]{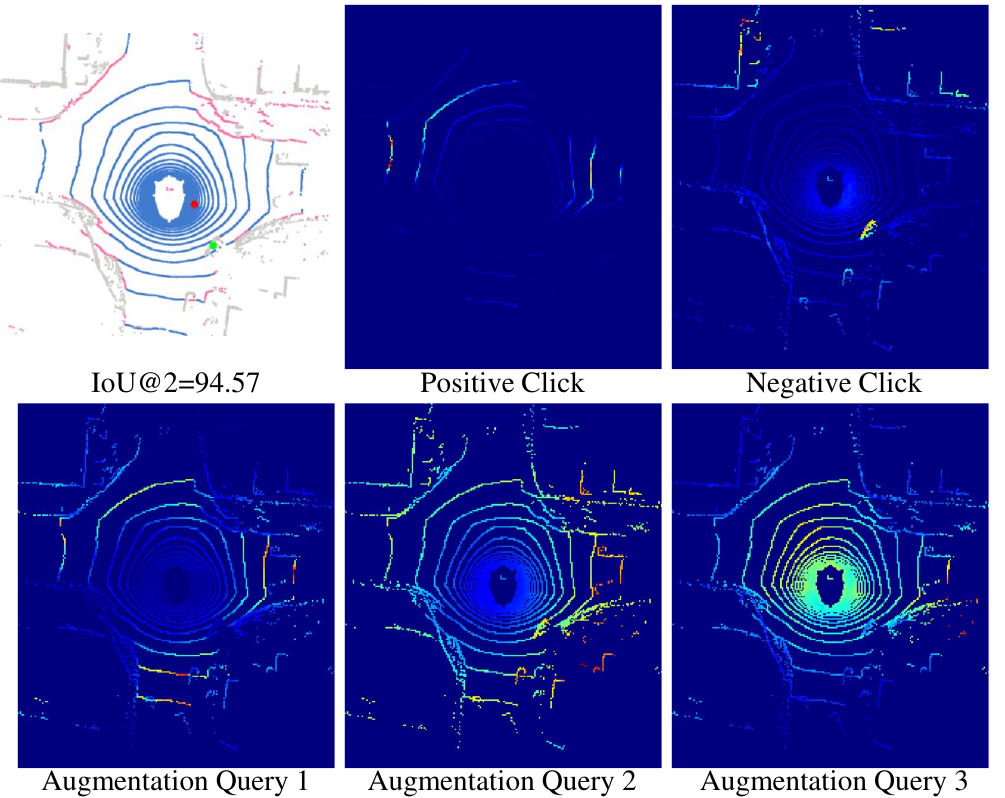}
    \caption{\textbf{Attention Map of Click Queries and Augmentation Queries.} Red points indicate positive clicks, and green points indicate negative clicks. For the point cloud segmentation results, blue represents true positives, pink represents false positives, yellow represents false negatives, and gray represents true negatives. All visualization results follow the same color scheme. \vspace{-3.5mm}
    }
    \label{fig:query_aug}
    \vspace{-4mm}
\end{figure}

Fig \ref{fig:query_aug} effectively illustrates the role of the query augmentation module. In the segmentation process for the road category, although the positive click query only focused on a small portion of the road, the augmentation queries effectively supplemented attention to other parts of the road, specifically targeting the edges of the road, the middle area, and the inner circle. With the assistance of the query augmentation module, the segmentation model complete the segmentation of the large-scale instance, the road, with only 2 user clicks, achieving a high accuracy of 94.57\% IoU, effectively overcoming the impact of scale disparity.

Each query is modeled as follows:
\begin{equation}\label{eq:query}
    q_k=(c_k, pe_k)=\begin{cases}
    (v_k+v_{pos}, pe_k) & \text{for positive clicks,} \\
    (v_k+v_{neg}, pe_k) & \text{for negative clicks,} \\
    (v_k, pe_k) & \text{for augmentation points.} \\
    \end{cases}
\end{equation}
The query $q_k$ is composed of two parts $(c_k, pe_k)$, separately encoding the content and position of the query. The position part $pe_k$ converts the point coordinate through Fourier positional embedding, and the content part $v_{pos}$ is a learnable vector characterizing positive clicks and $v_{neg}$ for negative clicks, so as to differentiate them from augmentation queries. $v_k$ is a learnable vector that extracts the mask embedding corresponding to the query through the attention mechanism in the subsequent query-voxel transformer. $v_k$ is all initialized as 0, thereby emphasizing the role of the position part in the attention mechanism during the initial stage.

\subsection{Mask Decoder}

\textbf{Two-way Query-voxel Transformer.} The query-voxel transformer enables the queries and voxel embeddings to update each other through a bidirectional attention mechanism.  Each transformer layer performs 4 steps: 1) self-attention among the queries (Q2Q), 2) cross-attention from queries to voxel embeddings (Q2V), 3) a MLP updates each query, and 4) inverse cross-attention from voxel embeddings to queries (V2Q). 

The Q2Q self-attention allows augmentation queries to gain prompts about the desired mask from click queries, while also performing information exchange among different queries. The Q2V cross-attention enables queries to update its content part using voxel embeddings, thereby extracting features of the corresponding mask in the local neighborhood. Unlike the previous three steps, which only update the queries, the final V2Q cross-attention inversely utilizes the prompt information in the queries to update the voxel embeddings. This makes the voxel embeddings click-aware, enhancing the segmentation precision of the output mask within local regions, thereby improving the segmentation accuracy, especially for small-scale instances. 

\textbf{Global Attention in Transformer.} It is noteworthy that after introducing the query augmentation module, continuing to use the common local attention in general point cloud segmentation transformers makes the model more prone to false positives when segmenting small-scale instances. With an increased number of queries participating in the segmentation, if a query only focuses on its local neighborhood, it might mistakenly include instances that are far from the user's desired mask as part of the foreground, leading to the generation of false positives. To mitigate this adverse effect, we replace local attention with global attention in all attention layers. This adjustment allows each query to no longer just focus on its local neighborhood but to exchange information with all other queries and voxels. Consequently, this optimizes the segmentation mask on a global level and suppresses the generation of false positives. Due to our model utilizing significantly fewer queries than the general point cloud segmentation transformer (hundreds vs. tens of thousands), the additional computational cost brought by global attention is completely acceptable.

\begin{figure} [t]
    \centering
    \includegraphics[width=\linewidth]{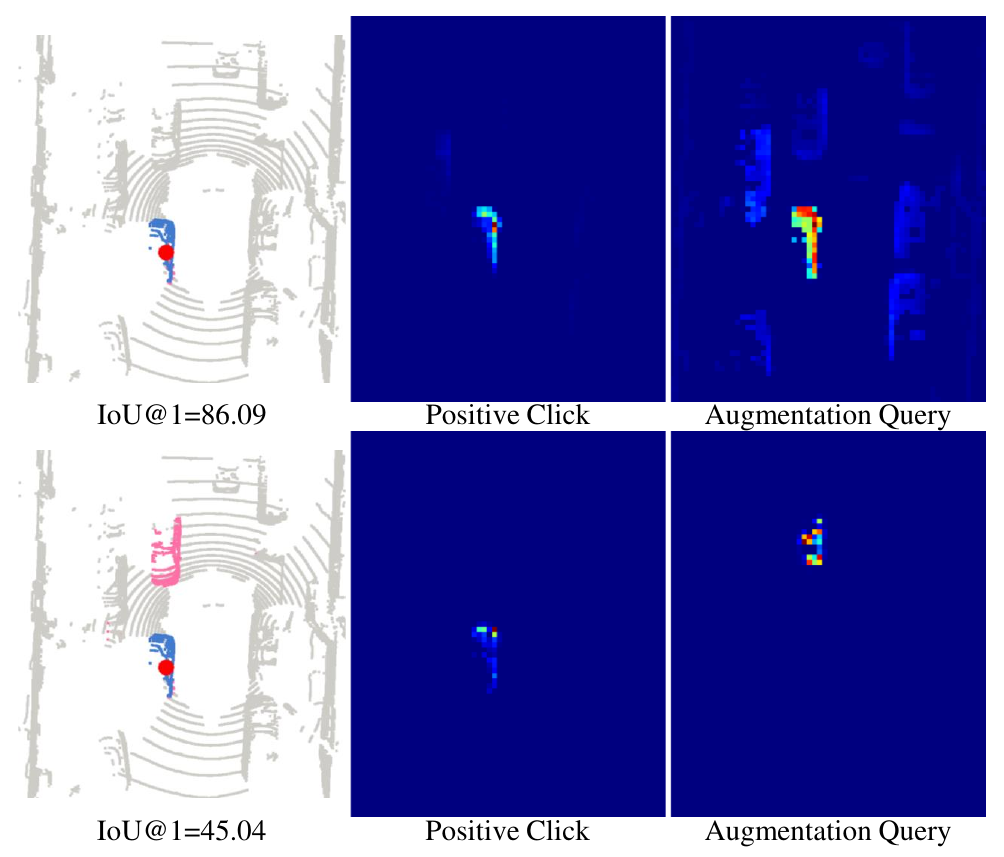}
    \caption{\textbf{Comparison of Global Attention (top) and Local Attention (bottom): Attention Maps and Segmentation Results.}\vspace{-3.5mm}
    }
    \label{fig:global_attn}
    \vspace{-4mm}
\end{figure}

Fig \ref{fig:global_attn} further illustrates the suppressive effect of global attention on false positives. When using local attention, due to the focus on the local neighborhood, a distant augmentation query mistakenly identifies a nearby instance of the same category (another car) as the desired instance, paying high attention to it, ultimately leading to a large number of false positives in the segmentation results. However, by switching to global attention, through the exchange of information between the query and all other queries and voxels, distant augmentation queries can also correctly focus on the desired instance. At the same time, they are able to recognize other instances of the same category (with attention higher than the background but lower than the desired instance), thereby suppressing the generation of false positives and nearly doubling the segmentation accuracy.

\textbf{Mask Segmentation Module.} To perform point-wise segmentation, we first need to extract the features $x_i$ of each point from the voxel features updated by the query-voxel transformer:
\begin{equation}\label{eq:pointfeature}
    x_i = f_{extractor}([x_{voxel}, pe_{rel}])
\end{equation}
where $x_{voxel}$ is the feature of the voxel to which the point belongs and $pe_{rel}$ encodes the relative position of the point within the voxel. Next, we calculate the probability of the $i$th point belonging to the $k$th binary mask based on the similarity between their features:
\begin{equation}\label{eq:binarymask}
    mask_{i, k}=\sigma(\langle x_i, c_k\rangle)
\end{equation}
where $\langle,\rangle$ represents the dot product, and $\sigma$ is the \emph{sigmoid} function to ensure the result is in $[0,1]$. Meanwhile, we predict the probability of the $k$th binary mask belongs to the foreground with the query content:
\begin{equation}\label{eq:maskprob}
    prob_k = f_{predictor}(c_k)
\end{equation}
Ultimately, we aggregate the prediction of all queries to calculate the probability of each point belonging to the foreground:
\begin{equation}\label{eq:finalseg}
score_i=\sum_{k} mask_{i, k} \cdot prob_k
\end{equation}

All queries contribute to the final segmentation mask, so that the entire process can be optimized through the backpropagated gradients. During inference, we obtain the binary segmentation mask by thresholding $\{score_i\}_{i=1}^N$ (set to 0.5).

\subsection{Training Details}

\textbf{Loss Function.} We supervise our network with binary cross-entropy loss between the predicted segmentation mask $M$ without thresholding and the groundtruth binary mask $M_{gt}$. The loss is defined as: 
\begin{equation}\label{eq:loss}
\mathcal{L}=w_{class}[\lambda_{fg}\mathcal{L}_{BCE}(M_{fg}, M_{gt})+\lambda_{bg}\mathcal{L}_{BCE}(M_{bg}, M_{gt})]
\end{equation}
Herein, $w_{class}$ is the class weight coefficient to balance the proportion of different classes in the training data, while $\lambda_{fg}$ and $\lambda_{bg}$ are the foreground and background weight coefficients respectively, utilized to balance the quantity difference between them. For supervision augmentation, we compute the binary cross-entropy loss for the segmentation masks outputted by each layer of the transformer during training. While, during inference, only the output of the last layer is utilized to get the final segmentation mask.

\textbf{User Click Simulation.} During the training stage, obtaining a large quantity of real human user clicks is difficult and not essential. Therefore, we simulate user clicks for training through random sampling, akin to similar methods in 2D domain\cite{liew2017regional, li2018interactive, jang2019interactive}. Without loss of generality, we sample a set of positive clicks uniformly at random from the desired foreground mask, and the same to negative clicks sampled from the background. Additionally, the desired mask and the number of click prompts are also determined randomly each time. The fully random strategy promotes our interactive model to generalize well across various user behaviors in an open-world setting.

%%%%%%%%%%%%%%%%%%%%%%%%%%%%%%%%%%%%%%%%%%%%%%%%%%%%%%%%%%%%%%%%%%%
\section{Experiments}
\label{sec:exp}

\begin{figure*} [t]
    \centering
    \includegraphics[width=\linewidth]{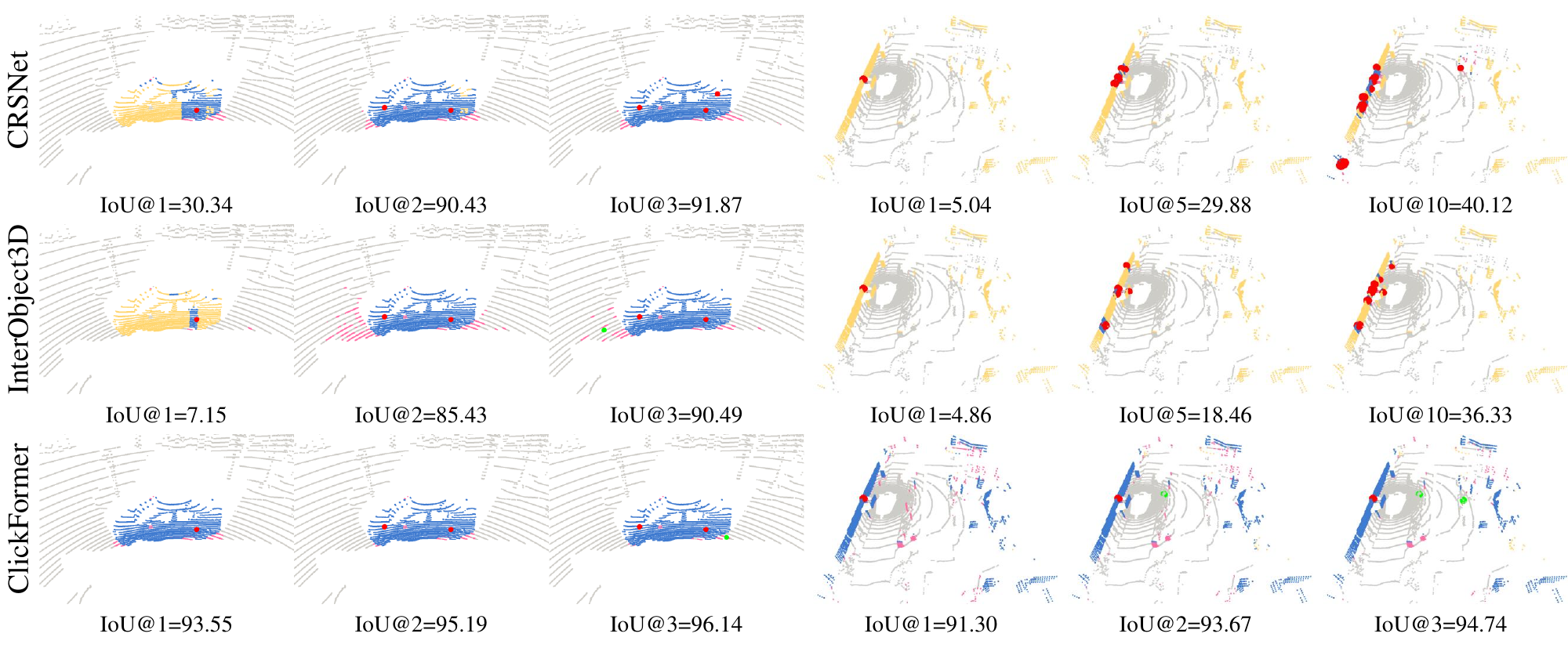}
    \caption{\textbf{Qualitative Results on Outdoor Datasets.} Red points indicate positive clicks, and green points indicate negative clicks. For the point cloud segmentation results, blue represents true positives, pink represents false positives, yellow represents false negatives, and gray represents true negatives. All visualization results follow the same color scheme. \vspace{-3.5mm}
    }
    \label{fig:outdoor}
    \vspace{-1mm}
\end{figure*}

\textbf{Datasets.} We leverage three key outdoor datasets for evaluating our segmentation model: nuScenes, SemanticKITTI, and KITTI360. nuScenes\cite{nuscenes} offers a diverse collection of urban environment data, crucial for autonomous driving research. SemanticKITTI\cite{behley2019iccv} extends the KITTI Vision Benchmark with dense annotations, enabling the study of temporal segmentation consistency. KITTI360\cite{Liao2022PAMI} further enriches this with high-resolution imagery and comprehensive semantic annotations, allowing for detailed urban scene analysis. These datasets collectively provide a robust framework for assessing our model's performance and generalization capabilities across different outdoor scenarios and varying conditions.

To assess our model's versatility in indoor environments, we utilized the ScanNet and S3DIS datasets as well. ScanNet\cite{dai2017scannet} provides a comprehensive collection of over 1500 scans of diverse indoor scenes reconstructed from RGB-D images, enriched with detailed semantic annotations for 3D segmentation. S3DIS\cite{armeni_cvpr16} offers extensive 3D scans of various indoor spaces, including offices, classrooms and so on, with precise semantic labels for numerous object classes. These datasets enable a thorough examination of our model's indoor segmentation capabilities, highlighting its adaptability to different settings.

\textbf{Baselines.} Given the limited research on interactive point cloud segmentation, we choose the only two existing interactive point cloud instance segmentation methods, CRSNet and InterObject3D, as our baselines. CRSNet\cite{sun2023click} encodes user clicks as an additional attribute of the point cloud and processes the segmentation masks through multiple MLPs. In contrast, InterObject3D\cite{kontogianni2023interactive} encodes user clicks into a 3D volume based on spatial relationships and obtains segmentation results through a sparse convolution-based 3D Unet.

\textbf{Evaluation Metrics.} We perform the evaluation by referencing the standard metric from the 2D domain\cite{benenson2019large, jang2019interactive, li2018interactive}, IoU@k, the average Intersection over Union (IoU) for k number of clicks per segmentation mask. Considering the significant variance in the number of instances across different categories, results averaged per instance are predominantly influenced by categories with a higher number of instances, which does not accurately reflect the segmentation model's ability to handle various categories. Therefore, we introduce \textbf{mIoU@k}, the IoU for k number of clicks per mask averaged across categories, ensuring equal contribution of all categories to the final metric, to more comprehensively assess the model's capability in segmenting different foreground categories.

\subsection{Outdoor Dataset Evaluation}

\begin{table*}[t]
\vspace{0.2cm}
\centering
\small
\caption{\textbf{Evaluation Metrics on Outdoor Datasets.}}
\begin{tabular}{@{}clcccccccc@{}}
\toprule
% row 1
\multirow{2}*{\textbf{Train$\rightarrow$ Eval}} &
\multirow{2}*{\textbf{Method}} 
& \multicolumn{2}{c}{\textbf{mIoU@1}} 
& \multicolumn{2}{c}{\textbf{mIoU@3}} 
& \multicolumn{2}{c}{\textbf{mIoU@5}} 
& \multicolumn{2}{c}{\textbf{mIoU@10}}
\\
\cmidrule(lr){3-4} \cmidrule(lr){5-6} \cmidrule(lr){7-8} \cmidrule(lr){9-10}
& & \emph{thing} & \emph{stuff} & \emph{thing} & \emph{stuff} & \emph{thing} & \emph{stuff} & \emph{thing} & \emph{stuff}\\
\midrule
\multirow{3}*{nuScenes$\rightarrow$ nuScenes} & CRSNet~\cite{sun2023click} & 31.24 & 7.05 & 43.91 & 17.03 & 46.39 & 22.67 & 50.13 & 31.75\\
&InterObject3D~\cite{kontogianni2023interactive} & 29.00 & 10.76 & 44.61 & 25.74 & 50.85 & 31.00 & 55.58 & 38.59\\
&\textbf{ClickFormer} & \textbf{35.19} & \textbf{48.58} & \textbf{50.21} & \textbf{62.19} & \textbf{56.90} & \textbf{64.32} & \textbf{60.56} & \textbf{65.37}\\

\midrule
\multirow{3}*{SemanticKITTI$\rightarrow$ KITTI360} & CRSNet~\cite{sun2023click} & 28.34 & 9.55 & 40.02 & 17.33 & 41.12 & 23.32 & 46.93 & 29.75\\
&InterObject3D~\cite{kontogianni2023interactive} & \textbf{33.99} & 12.25 & 42.61 & 23.06 & 45.80 & 29.61 & 49.61 & 36.11\\
&\textbf{ClickFormer} & 27.99 & \textbf{41.08} & \textbf{50.35} & \textbf{52.33} & \textbf{54.44} & \textbf{55.44} & \textbf{59.34} & \textbf{58.43}\\

\bottomrule
\end{tabular}
\label{tab:out_resu}
\vspace{-0.35cm}
\end{table*}

Results are summarized in Table \ref{tab:out_resu} and Fig \ref{fig:outdoor}. Firstly, we evaluate our method by training and testing within the same domain. This scenario is designed to test the capability of interactive segmentation methods to annotate the remaining raw data after being trained with a portion of labeled data, thereby expanding the annotated dataset. For a fair comparison, both our method and baseline methods utilize GD-MAE, pretrained on KITTI, as the encoder, and are retrained on the nuScenes training set under the same protocol, followed by testing on the validation set. Our method significantly outperforms both of the baselines with the same number of user clicks. For the thing categories, our method consistently maintains a lead of at least $\sim$5\% mIoU. As for the stuff categories, the advantage of our method is even more pronounced. Even for instances that are as long as tens of meters such as roads, or as large as an entire building, our method achieves 48.58\% mIoU with just 1 click, which the baselines cannot reach even with 10 clicks. Moreover, the performance can be further improved to 65.37\% mIoU with 10 clicks.

To further evaluate our method's performance on out-of-domain scenes, we retrain both our method and baseline methods on the SemanticKITTI dataset under the same protocol and then tested them on the KITTI360 dataset without finetuning. This scenario is designed to test the capability of interactive segmentation methods to annotate unseen scene data after being trained on existing public datasets, thereby generating brand-new datasets. Since the test set contains scenes completely different from the training set, this greatly tests the methods' generalization capabilities. Our method's performance in unseen scenes is close to that in trained scenes, achieving 59.34\% mIoU on thing categories and 58.43\% mIoU on stuff categories with 10 clicks, still significantly outperforming both of the baselines.

The experimental results demonstrate that our method effectively addresses the scale disparity of instances in interactive segmentation. As a result, it not only achieves high segmentation accuracy for both small-scale instances of thing categories and large-scale instances of stuff categories, but also generalizes well to unseen scenes. These aspects are crucial for the application of interactive segmentation methods.

\begin{figure*} [t]
    \centering
    \includegraphics[width=\linewidth]{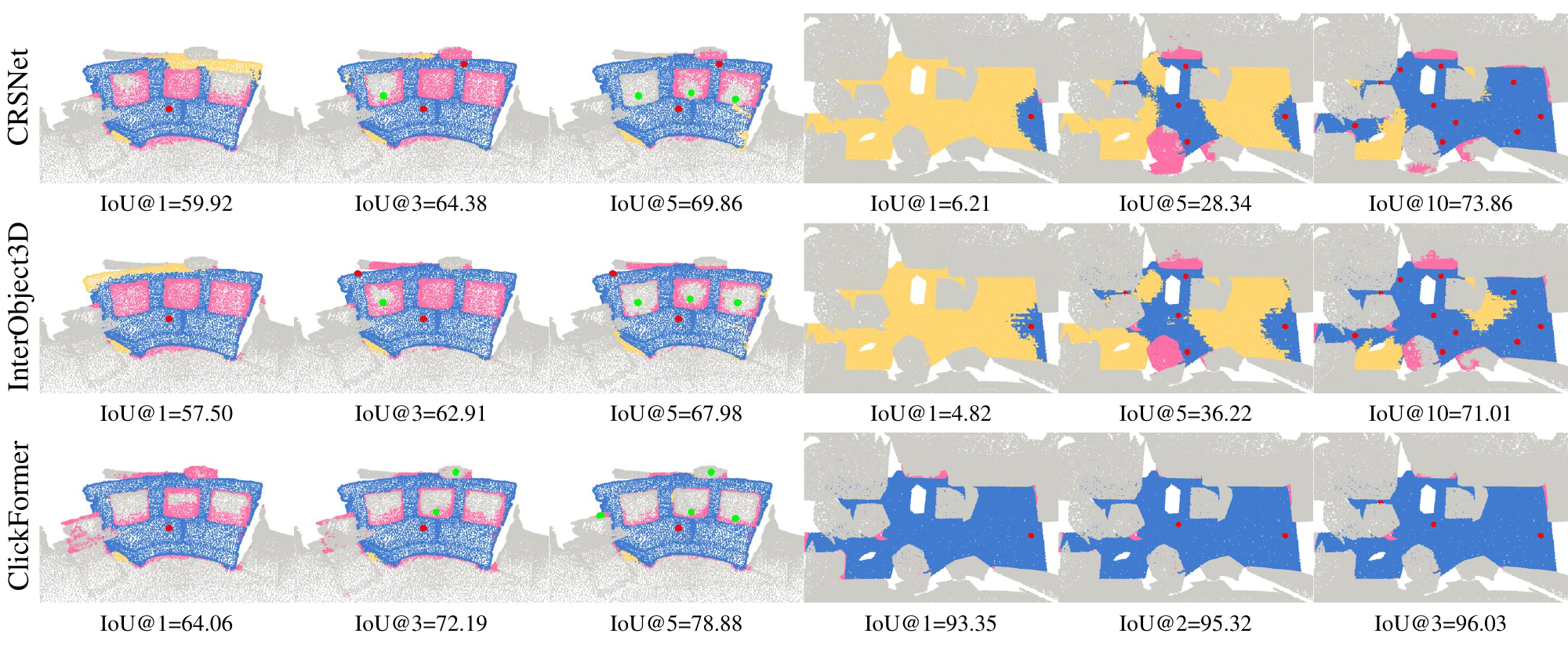}
    \caption{\textbf{Qualitative Results on Indoor Datasets.}\vspace{-3.5mm}
    }
    \label{fig:indoor}
    \vspace{-1mm}
\end{figure*}

\subsection{Indoor Dataset Evaluation}

\begin{table*}[t]
\vspace{0.2cm}
\centering
\small
\caption{\textbf{Evaluation Metrics on Indoor Datasets.}}
\begin{tabular}{@{}clcccccccc@{}}
\toprule
% row 1
\multirow{2}*{\textbf{Train$\rightarrow$ Eval}} &
\multirow{2}*{\textbf{Method}} 
& \multicolumn{2}{c}{\textbf{mIoU@1}} 
& \multicolumn{2}{c}{\textbf{mIoU@3}} 
& \multicolumn{2}{c}{\textbf{mIoU@5}} 
& \multicolumn{2}{c}{\textbf{mIoU@10}}
\\
\cmidrule(lr){3-4} \cmidrule(lr){5-6} \cmidrule(lr){7-8} \cmidrule(lr){9-10}
& & \emph{thing} & \emph{stuff} & \emph{thing} & \emph{stuff} & \emph{thing} & \emph{stuff} & \emph{thing} & \emph{stuff}\\
\midrule
\multirow{3}*{ScanNet$\rightarrow$ ScanNet} & CRSNet~\cite{sun2023click} & 35.05 & 11.68 & 46.18 & 24.02 & 48.62 & 29.96 & 49.88 & 36.41\\
&InterObject3D~\cite{kontogianni2023interactive} & 29.09 & 8.90 & 44.11 & 22.42 & 48.30 & 29.94 & 50.24 & 37.81\\
&\textbf{ClickFormer} & \textbf{40.15} & \textbf{52.60} & \textbf{50.23} & \textbf{63.89} & \textbf{54.05} & \textbf{67.54} & \textbf{56.46} & \textbf{70.00}\\

\midrule
\multirow{3}*{ScanNet$\rightarrow$ S3DIS} & CRSNet~\cite{sun2023click} & 30.05 & 12.16 & 48.75 & 26.00 & 53.87 & 36.30 & 57.55 & 47.53\\
&InterObject3D~\cite{kontogianni2023interactive} & 21.59 & 8.79 & 42.06 & 18.93 & 49.54 & 27.92 & 55.14 & 40.40\\
&\textbf{ClickFormer} & \textbf{43.36} & \textbf{59.69} & \textbf{54.25} & \textbf{73.89} & \textbf{58.09} & \textbf{76.89} & \textbf{61.59} & \textbf{80.29}\\

\bottomrule
\end{tabular}
\label{tab:in_resu}
\vspace{-0.35cm}
\end{table*}

Results are summarized in Table \ref{tab:in_resu} and Fig \ref{fig:indoor}. We evaluate our method on the ScanNet dataset to assess its performance in indoor scenes within the same domain. This scenario is designed to test the capability of interactive segmentation methods to handle different types of point cloud obtained from various environments and sensors. Considering the point cloud input channels and scene sizes suitable for feature extractor, we choose the pretrained MinkowskiUNet as the encoder for all methods. Although the scale disparity of instances in indoor scenes is somewhat not as severe as in outdoor scenes, the two baseline methods still perform poorly in segmenting the stuff categories, failing to reach 40\% mIoU even with 10 clicks. In contrast, our method achieves 52.60\% mIoU with only 1 click and further improved to 70.00\% mIoU with 10 clicks. Simultaneously, our method consistently outperforms in segmenting the thing categories, achieving approximately 6\% higher mIoU than the two baseline methods with 5 or 10 clicks.

Similar to the experimental setup on outdoor datasets, we test the model retrained on the ScanNet dataset on the S3DIS dataset without finetuning, to evaluate the performance of our method in out-of-domain scenarios. Given that the scenes in the test set are completely different and contain many unseen categories, this poses a significant challenge to the generalization ability of the interactive segmentation models. All methods tested on the S3DIS dataset show even slightly higher performance metrics than ScanNet. Similar observations have been made in other works involving this set of transfer experiments, demonstrating that the differences in dataset such as quality and evaluation categories are responsible for this abnormal phenomenon. Nevertheless, our method achieves higher segmentation accuracy over the baseline methods for both thing and stuff categories even with fewer clicks, and the accuracy significantly improves as the number of clicks increases, reaching 61.59\% and 80.29\% mIoU respectively with 10 clicks.

The experimental results demonstrate that our method can adapt to different encoders, thereby handling different types of point cloud data obtained from various environments and sensors. Our approach is capable of enhancing the segmentation accuracy of instances at diverse scales, while also effectively dealing with novel scenes and unseen category, without the need for additional training.

\subsection{Ablation Studies}

We ablated parts of our model architecture to validate the effectiveness and rationality of our method, and the results are summarized in Table \ref{tab:abl_resu}. All ablation experiments are conducted on the nuScenes dataset.

\begin{table}[t]
\vspace{0.2cm}
\centering
\small
\caption{\textbf{Evaluation Metrics of Ablation Experiments.}}
\begin{tabular}{@{}lcc@{}}
\toprule
% row 1
\multirow{2}*{\textbf{Method}} 
& \multicolumn{2}{c}{\textbf{mIoU@10}}
\\
\cmidrule(lr){2-3} 
& \emph{thing} & \emph{stuff}\\
\midrule
\textbf{ClickFormer} & 60.56 & 65.37 \\
\midrule
w/o augmentation queries & 47.79 & 49.97 \\
w/o V2Q cross-attention & 35.46 & 60.18 \\
w/o global attention & 50.62 & 60.22 \\
\midrule
w/ 2$\times$ augmentation queries & 62.23 & 67.65 \\
\bottomrule
\end{tabular}
\label{tab:abl_resu}
\vspace{-0.35cm}
\end{table}

\textbf{Augmentation Queries.} Firstly, we evaluate a model that only uses click queries, with the rest of the network architecture remaining unchanged. Without augmentation queries, both thing and stuff categories show a significant decrease in segmentation accuracy, approximately 15\% mIoU. This demonstrates that augmentation queries play an irreplaceable role in addressing the scale disparity of instances. By query augmentation module, not only is the model's understanding of global scene information enhanced, but it also ensures that the attention scope of each query is independent of instance scales. This effectively mitigates the impact of scale disparity during model training, benefiting the segmentation of both thing and stuff categories.

Furthermore, we evaluated a model that uses double the number of exploratory queries. With the increase in the number of augmentation queries, both thing and stuff categories see a slight improvement in segmentation accuracy. However, a greater number of queries means higher computational costs and memory overhead, indicating a trade-off between cost and accuracy. Considering the diminishing marginal utility of accuracy improvements, using an excessive number of augmentation queries is not cost-effective.

\textbf{Two-way Attention.} In our model, the query-voxel transformer incorporates an inverse cross-attention mechanism from voxel embeddings to queries, which is an enhancement over the vanilla transformer. Without the V2Q cross-attention, the segmentation accuracy of thing categories experiences a significant decrease of about 25\% mIoU. The rationale behind this is that the variance between different instance features of thing categories is much greater than that of stuff categories. The V2Q cross-attention leverages prompt information within queries to update voxel embeddings, making them click-aware and more similar to the corresponding query features. It improves the effectiveness of the subsequent mask segmentation module.

\textbf{Global Attention.} When using local attention in place of the global attention in our model, the decrease in segmentation accuracy of thing categories is about twice that of stuff categories. This indicates that global attention is more crucial for thing categories. While the query augmentation module makes the model more likely to generate false positives when segmenting small-scale instances, the global attention allows for comprehensive information exchange between queries and the global point cloud, effectively mitigating the negative impact of false positives on segmentation accuracy.

\begin{figure} [h]
    \centering
    \includegraphics[width=\linewidth]{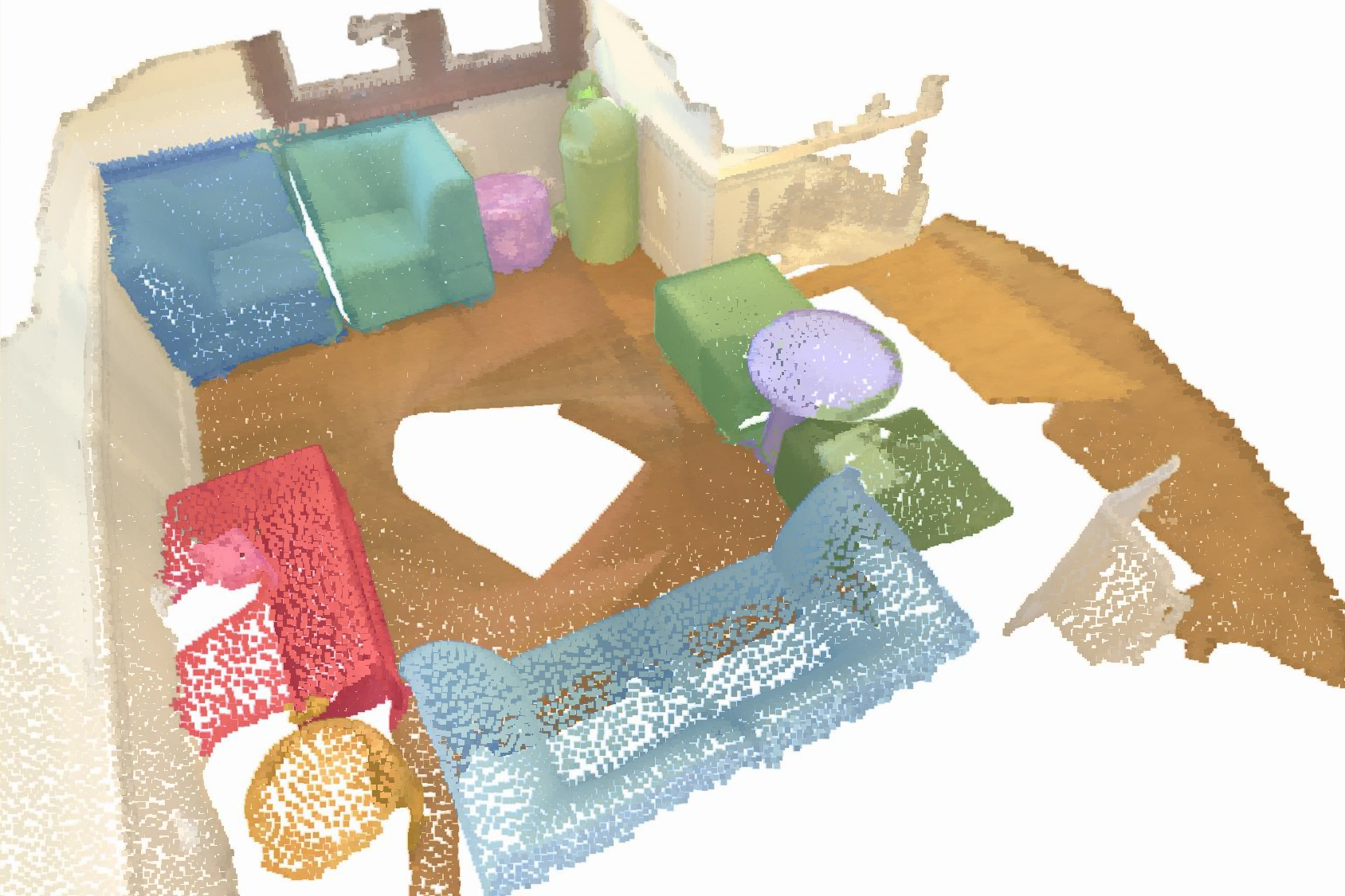}
    \caption{\textbf{Visualization Result of Case Study.}\vspace{-3.5mm}
    }
    \label{fig:multi}
    \vspace{-2mm}
\end{figure}

\subsection{Case Study}

To demonstrate the application of our method, we annotated a scene from ScanNet with 16 instances (including stuff categories). With only 1.88 clicks and 0.09s inference time per instance, ClickFormer achieved 76.08\% mIoU, and the visulization results are shown in Fig. \ref{fig:multi}.

%%%%%%%%%%%%%%%%%%%%%%%%%%%%%%%%%%%%%%%%%%%%%%%%%%%%%%%%%%%%%%%%%%%
\section{Conclusion}
\label{sec:conclude}

In this paper, we first propose the issue of scale disparity of instances in interactive point cloud segmentation. Through the query augmentation module, our interactive segmentation model effectively resolves the scale disparity, achieving significant improvements in segmentation accuracy, especially for large-scale instances. We further enhance segmentation performance by adopting global attention and other network structure improvements. Our method demonstrates good generalization across indoor and outdoor environments, novel scenes, and unseen categories. We believe our work will inspire further research in interactive point cloud segmentation.


%%%%%%%%%%%%%%%%%%%%%%%%%%%%%%%%%%%%%%%%%%%%%%%%%%%%%%%%%%%%%%%%%%%%%%%%%%%%%%%%

%%%%%%%%%%%%%%%%%%%%%%%%%%%%%%%%%%%%%%%%%%%%%%%%%%%%%%%%%%%%%%%%%%%%%%%%%%%%%%%%


% \section*{ACKNOWLEDGMENT}

%%%%%%%%%%%%%%%%%%%%%%%%%%%%%%%%%%%%%%%%%%%%%%%%%%%%%%%%%%%%%%%%%%%%%%%%%%%%%%%%

\bibliographystyle{IEEEtran}
\bibliography{root}

\begin{thebibliography}{10}
\providecommand{\url}[1]{#1}
\csname url@samestyle\endcsname
\providecommand{\newblock}{\relax}
\providecommand{\bibinfo}[2]{#2}
\providecommand{\BIBentrySTDinterwordspacing}{\spaceskip=0pt\relax}
\providecommand{\BIBentryALTinterwordstretchfactor}{4}
\providecommand{\BIBentryALTinterwordspacing}{\spaceskip=\fontdimen2\font plus
\BIBentryALTinterwordstretchfactor\fontdimen3\font minus \fontdimen4\font\relax}
\providecommand{\BIBforeignlanguage}[2]{{%
\expandafter\ifx\csname l@#1\endcsname\relax
\typeout{** WARNING: IEEEtran.bst: No hyphenation pattern has been}%
\typeout{** loaded for the language `#1'. Using the pattern for}%
\typeout{** the default language instead.}%
\else
\language=\csname l@#1\endcsname
\fi
#2}}
\providecommand{\BIBdecl}{\relax}
\BIBdecl

\bibitem{sun2023click}
W.~Sun, Z.~Luo, Y.~Chen, H.~Li, J.~Marcato, W.~N. Gon{\c{c}}alves, and J.~Li, ``A click-based interactive segmentation network for point clouds,'' \emph{IEEE Transactions on Geoscience and Remote Sensing}, 2023.

\bibitem{kontogianni2023interactive}
T.~Kontogianni, E.~Celikkan, S.~Tang, and K.~Schindler, ``Interactive object segmentation in 3d point clouds,'' in \emph{2023 IEEE International Conference on Robotics and Automation (ICRA)}.\hskip 1em plus 0.5em minus 0.4em\relax IEEE, 2023, pp. 2891--2897.

\bibitem{yue2023agile3d}
Y.~Yue, S.~Mahadevan, J.~Schult, F.~Engelmann, B.~Leibe, K.~Schindler, and T.~Kontogianni, ``Agile3d: Attention guided interactive multi-object 3d segmentation,'' \emph{arXiv preprint arXiv:2306.00977}, 2023.

\bibitem{gao2020we}
B.~Gao, Y.~Pan, C.~Li, S.~Geng, and H.~Zhao, ``Are we hungry for 3d lidar data for semantic segmentation? a survey and experimental study,'' \emph{arXiv preprint arXiv:2006.04307}, 2020.

\bibitem{xu2020weakly}
X.~Xu and G.~H. Lee, ``Weakly supervised semantic point cloud segmentation: Towards 10x fewer labels,'' in \emph{Proceedings of the IEEE/CVF conference on computer vision and pattern recognition}, 2020, pp. 13\,706--13\,715.

\bibitem{unal2022scribble}
O.~Unal, D.~Dai, and L.~Van~Gool, ``Scribble-supervised lidar semantic segmentation,'' in \emph{Proceedings of the IEEE/CVF Conference on Computer Vision and Pattern Recognition}, 2022, pp. 2697--2707.

\bibitem{sofiiuk2022reviving}
K.~Sofiiuk, I.~A. Petrov, and A.~Konushin, ``Reviving iterative training with mask guidance for interactive segmentation,'' in \emph{2022 IEEE International Conference on Image Processing (ICIP)}.\hskip 1em plus 0.5em minus 0.4em\relax IEEE, 2022, pp. 3141--3145.

\bibitem{chen2022focalclick}
X.~Chen, Z.~Zhao, Y.~Zhang, M.~Duan, D.~Qi, and H.~Zhao, ``Focalclick: Towards practical interactive image segmentation,'' in \emph{Proceedings of the IEEE/CVF Conference on Computer Vision and Pattern Recognition}, 2022, pp. 1300--1309.

\bibitem{liu2023simpleclick}
Q.~Liu, Z.~Xu, G.~Bertasius, and M.~Niethammer, ``Simpleclick: Interactive image segmentation with simple vision transformers,'' in \emph{Proceedings of the IEEE/CVF International Conference on Computer Vision}, 2023, pp. 22\,290--22\,300.

\bibitem{du2023efficient}
F.~Du, J.~Yuan, Z.~Wang, and F.~Wang, ``Efficient mask correction for click-based interactive image segmentation,'' in \emph{Proceedings of the IEEE/CVF Conference on Computer Vision and Pattern Recognition}, 2023, pp. 22\,773--22\,782.

\bibitem{kirillov2023segment}
A.~Kirillov, E.~Mintun, N.~Ravi, H.~Mao, C.~Rolland, L.~Gustafson, T.~Xiao, S.~Whitehead, A.~C. Berg, W.-Y. Lo \emph{et~al.}, ``Segment anything,'' \emph{arXiv preprint arXiv:2304.02643}, 2023.

\bibitem{shen2020interactive}
T.~Shen, J.~Gao, A.~Kar, and S.~Fidler, ``Interactive annotation of 3d object geometry using 2d scribbles,'' in \emph{Computer Vision--ECCV 2020: 16th European Conference, Glasgow, UK, August 23--28, 2020, Proceedings, Part XVII 16}.\hskip 1em plus 0.5em minus 0.4em\relax Springer, 2020, pp. 751--767.

\bibitem{zhi2022ilabel}
S.~Zhi, E.~Sucar, A.~Mouton, I.~Haughton, T.~Laidlow, and A.~J. Davison, ``ilabel: Revealing objects in neural fields,'' \emph{IEEE Robotics and Automation Letters}, vol.~8, no.~2, pp. 832--839, 2022.

\bibitem{goel2023interactive}
R.~Goel, D.~Sirikonda, S.~Saini, and P.~Narayanan, ``Interactive segmentation of radiance fields,'' in \emph{Proceedings of the IEEE/CVF Conference on Computer Vision and Pattern Recognition}, 2023, pp. 4201--4211.

\bibitem{yang2023gd}
H.~Yang, T.~He, J.~Liu, H.~Chen, B.~Wu, B.~Lin, X.~He, and W.~Ouyang, ``Gd-mae: generative decoder for mae pre-training on lidar point clouds,'' in \emph{Proceedings of the IEEE/CVF Conference on Computer Vision and Pattern Recognition}, 2023, pp. 9403--9414.

\bibitem{choy20194d}
C.~Choy, J.~Gwak, and S.~Savarese, ``4d spatio-temporal convnets: Minkowski convolutional neural networks,'' in \emph{Proceedings of the IEEE/CVF conference on computer vision and pattern recognition}, 2019, pp. 3075--3084.

\bibitem{liew2017regional}
J.~Liew, Y.~Wei, W.~Xiong, S.-H. Ong, and J.~Feng, ``Regional interactive image segmentation networks,'' in \emph{2017 IEEE international conference on computer vision (ICCV)}.\hskip 1em plus 0.5em minus 0.4em\relax IEEE, 2017, pp. 2746--2754.

\bibitem{li2018interactive}
Z.~Li, Q.~Chen, and V.~Koltun, ``Interactive image segmentation with latent diversity,'' in \emph{Proceedings of the IEEE Conference on Computer Vision and Pattern Recognition}, 2018, pp. 577--585.

\bibitem{jang2019interactive}
W.-D. Jang and C.-S. Kim, ``Interactive image segmentation via backpropagating refinement scheme,'' in \emph{Proceedings of the IEEE/CVF Conference on Computer Vision and Pattern Recognition}, 2019, pp. 5297--5306.

\bibitem{nuscenes}
H.~Caesar, V.~Bankiti, A.~H. Lang, S.~Vora, V.~E. Liong, Q.~Xu, A.~Krishnan, Y.~Pan, G.~Baldan, and O.~Beijbom, ``nuscenes: A multimodal dataset for autonomous driving,'' in \emph{CVPR}, 2020.

\bibitem{behley2019iccv}
J.~Behley, M.~Garbade, A.~Milioto, J.~Quenzel, S.~Behnke, C.~Stachniss, and J.~Gall, ``{SemanticKITTI: A Dataset for Semantic Scene Understanding of LiDAR Sequences},'' in \emph{Proc. of the IEEE/CVF International Conf.~on Computer Vision (ICCV)}, 2019.

\bibitem{Liao2022PAMI}
Y.~Liao, J.~Xie, and A.~Geiger, ``{KITTI}-360: A novel dataset and benchmarks for urban scene understanding in 2d and 3d,'' \emph{Pattern Analysis and Machine Intelligence (PAMI)}, 2022.

\bibitem{dai2017scannet}
A.~Dai, A.~X. Chang, M.~Savva, M.~Halber, T.~Funkhouser, and M.~Nie{\ss}ner, ``Scannet: Richly-annotated 3d reconstructions of indoor scenes,'' in \emph{Proc. Computer Vision and Pattern Recognition (CVPR), IEEE}, 2017.

\bibitem{armeni_cvpr16}
I.~Armeni, O.~Sener, A.~R. Zamir, H.~Jiang, I.~Brilakis, M.~Fischer, and S.~Savarese, ``3d semantic parsing of large-scale indoor spaces,'' in \emph{Proceedings of the IEEE International Conference on Computer Vision and Pattern Recognition}, 2016.

\bibitem{benenson2019large}
R.~Benenson, S.~Popov, and V.~Ferrari, ``Large-scale interactive object segmentation with human annotators,'' in \emph{Proceedings of the IEEE/CVF conference on computer vision and pattern recognition}, 2019, pp. 11\,700--11\,709.

\end{thebibliography}

\clearpage

\end{document}